\documentclass[preprint,12pt,3p]{elsarticle}

\usepackage{algorithm,algorithmic}
\usepackage{multirow}
\usepackage{hhline}
\usepackage{color,soul}
\usepackage{graphicx,subfigure}
\usepackage{tabularx}
\usepackage{booktabs}
\usepackage{tabu}
\usepackage{lscape}
\usepackage[geometry]{ifsym}
\usepackage{amsmath}

\let\oldtabular\tabular 
\renewcommand{\tabular}{\scriptsize\oldtabular}

\usepackage{amssymb}

\begin{document}

\begin{frontmatter}

\title{Covariate Shift Estimation based Adaptive Ensemble Learning for Handling Non-Stationarity in Motor Imagery related EEG-based Brain-Computer Interface}

\author[label1]{Haider Raza\corref{cor1}}

\address[label1]{School of Computer Science and Electronic Engineering, University of Essex, Colchester, UK}
\address[label2]{School of Computing and Intelligent Systems, Ulster University, Magee campus, Derry~Londonderry, UK}
\address[label3]{The Farr Institute of Health Informatics Research, Swansea University, Swansea, UK}
\address[label4]{Department of Computer Science, California State University Fresno, Fresno, CA, USA}

\cortext[cor1]{Haider Raza and Shang-Ming Zhou are the joint corresponding authors}
\ead{h.raza@essex.ac.uk}

\author[label2]{Dheeraj Rathee}
\ead{rathee-d@email.ulster.ac.uk}

\author[label3]{Shang-Ming Zhou\corref{cor1}}
\ead{s.zhou@swansea.ac.uk}

\author[label4]{Hubert Cecotti}
\ead{hcecotti@csufresno.edu}

\author[label2]{Girijesh Prasad}
\ead{g.prasad@ulster.ac.uk}

\begin{abstract}
The non-stationary nature of electroencephalography (EEG) signals makes an EEG-based brain-computer interface (BCI) a dynamic system, thus improving its performance is a challenging task. In addition, it is well-known that due to non-stationarity based covariate shifts, the input data distributions of EEG-based BCI systems change during inter- and intra-session transitions, which poses great difficulty for developments of online adaptive data-driven systems. Ensemble learning approaches have been used previously to tackle this challenge. However, passive scheme based implementation leads to poor efficiency while increasing high computational cost. This paper presents a novel integration of covariate shift estimation and unsupervised adaptive ensemble learning (CSE-UAEL) to tackle non-stationarity in motor-imagery (MI) related EEG classification. The proposed method first employs an exponentially weighted moving average model to detect the covariate shifts in the common spatial pattern features extracted from MI related brain responses. Then, a classifier ensemble was created and updated over time to account for changes in streaming input data distribution wherein new classifiers are added to the ensemble in accordance with estimated shifts. Furthermore, using two publicly available BCI-related EEG datasets, the proposed method was extensively compared with the state-of-the-art single-classifier based passive scheme, single-classifier based active scheme and ensemble based passive schemes. The experimental results show that the proposed active scheme based ensemble learning algorithm significantly enhances the BCI performance in MI classifications.
\end{abstract}


\end{frontmatter}


\section*{Acronyms}
\begin{flushleft}
BCI: Brain-computer-interface\\
CS: Covariate shift\\
CSP: Common spatial pattern\\
CSA: Covariate shift adaptation\\
CSE: Covariate shift estimation\\
CSE-UAEL: CSE-based unsupervised adaptive ensemble learning\\
CSV: Covariate shift validation\\
CSW: Covariate shift warning\\
DWEC: Dynamically weighted ensemble classification\\
EEG: Electroencephalography\\
ERD: Synchronization\\
ERS: Desynchronization\\
FB: Frequency band\\
FBCSP: Filter bank common spatial pattern\\
EWMA: exponential weighted moving average\\
$K$NN: $K$-nearest-neighbors\\
LDA: Linear discriminant analysis\\
MI: Motor imagery\\
NSL: Non-stationary learning\\
PCA: Principal component analysis\\
PW$K$NN: Probabilistic weighted $K$-nearest neighbour\\
RSM: Random subspace method\\
SSL: Semi-supervised learning\\
\end{flushleft}
\section{Introduction}
\label{sec1_intro}

\noindent Streaming data analytics has increasingly become the bedrock in many domains, such as bio-medical sciences, healthcare, and financial services. However, the majority of streaming data systems assume that the distributions of streaming data do not change over time. In reality, the streaming data obtained from real-world systems often possess non-stationary characteristics \cite{Alippi2008}. Such systems are often characterized by continuous evolving natures and thus, their behaviours often shift over time due to thermal drifts, aging effects, or other non-stationary environmental factors etc. These characteristics can adversely affect environmental, natural, artificial and industrial processes \cite{Ditzler2015a}. Hence, adaptive learning in a non-stationary environment (NSE), wherein the input data distribution shifts over time, is a challenging task. Developing machine learning models that can be optimized for non-stationary environments is in high demand. Currently machine learning methods for non-stationary systems are majorly categorized into passive and active approaches \cite{Ditzler2015a}. In the passive approach to non-stationary learning (NSL), it is assumed that the input distribution should be continuously shifting over time \cite{Alippi2013, Ditzler2015a}. Thus, passive scheme based methods adapt to new data distributions continuously for each new incoming observation or a new batch of observations from the streaming data. In contrast, an active scheme based NSL method uses a shift detection test to detect the presence of shifts in the streaming data, and an adaptive action is initiated based upon the time of detected shift\cite{Alippi2008a}. There exits a range of literature on transfer learning and domain adaptation theory, which aims to adapt to NSEs by transferring knowledge between training and test domains. In this case, one can match the features distribution of training and testing by the density ratio estimation approaches such as kernel mean matching \cite{Pan2010}, Kullback-Leibler importance estimation procedure, and least-squares importance fitting \cite{Sugiyama2007}. In addition to density ratio estimation methods, several methods, such as domain adaption with conditional transferable components, try to minimize the domain shift by finding invariant representation across training and target domains \cite{Gong2016}. In fact, to favorably transfer knowledge between domains, one needs to estimate the primary causal mechanism of the data generating process. These methods have, however, a limited applicability in real world problems, where the data in test domain are generated while operating in real-time. \\

A typical brain-computer-interface (BCI) system aims to provide an alternative means of communication or rehabilitation for the physically challenged population so as to allow them to express their wills without muscle exertion \cite{Wolpaw2002}. An electroencephalography (EEG)-based BCI is such a non-stationary system \cite{Zhou2008} and quasi-stationary segment in EEG signals have duration of nearly 0.25 sec \cite{Celka2005}.  The non-stationarities of the EEG signals may be caused by various events, such as changes in the user attention levels, electrode placements, or user fatigues \cite{Li2010a, Arvaneh2013a, Raza2015_PR}. In other words, the basic cause of the non-stationarity in EEG signals is not only associated with the influences of the external stimuli to the brain mechanisms, but the switching of the cognitive task related inherent metastable states of neural assemblies also contributes towards it \cite{Rathee2017}. These non-stationarities cause notable variations or shifts in the EEG signals both during trial-to-trial, and session-to-session transfers \cite{Blankertz2008b,Raza2015_PR,Raza2016_SC,Chowdhury2018}. As a result, these variations often appear as covariate shifts (CSs) wherein the input data distributions differ significantly between training and testing phases while the conditional distribution remains the same \cite{Sugiyama2007,Raza2013b,Raza2013c,Raza2014,Raza2016PhD}. \\

Non-invasive EEG-based BCI systems acquire neural signals at scalp level to be analysed for evaluating activity-specific features of EEG signals e.g. voluntary imagery/execution tasks,  and finally the output signals are relayed to different control devices \cite{Wolpaw2002}. The EEG signals are acquired through a multichannel EEG amplifier, and a pre-processing step is performed to remove noise and enhance the signal-to-noise ratio. Then the discriminable features are extracted from the artefact-cleaned signals using feature extraction techniques, such as spatial filtering (e.g., common spatial pattern (CSP)) \cite{Ramoser2000}. Such a system operates typically in two phases, namely the training phase and the testing phase \cite{Lotte2007}. However, due to the non-stationary nature of the brain response characteristics, it is difficult to accurately classify the EEG patterns in motor imagery (MI) related BCI systems using traditional inductive algorithms \cite{Shenoy2006,Lotte2007}. For EEG-based BCI systems that operate online under real-time non-stationary/changing environments, it is required to consider the input features that are invariant to dataset shifts, or the learning approaches that can track the changes repeating over time, and the learning function can be adapted in a timely fashion. However, the traditional BCI systems are built upon passive approach to NSL for EEG signals. In passive schemes, both single and ensemble classifiers have been developed to improve the MI classification performance. In contrast, an active scheme based NSL in BCI systems provide a new option by estimating CSs in the streaming EEG features, in which an adaptive action can be initiated once the CS is confirmed. Our previous studies have demonstrated that the active approach to single-trial EEG classification outperformed existing passive approaches based BCI system \cite{Vidaurre2006,Shenoy2006,Satti2010,Li2010a,Liyanage2013,Raza2015}.\\

The aim of this paper is to extend our previous work and present a novel active scheme based unsupervised adaptive ensemble learning algorithm to adapt to CSs under non-stationary environments in EEG-based BCI systems. Different from the existing passive scheme based methods, the proposed algorithm is an active ensemble learning approach under non-stationary environments wherein a CS estimation test is used to detect at which point an updated classifier needs to be added to the ensemble during the evaluation phase. The transductive learning is implemented to enrich the training dataset during the evaluation phase using a probabilistic weighted $K$ nearest neighbour (PW$K$NN) method. Thus, a new classifier is added to the ensemble only when it is necessary, i.e. once the data from a novel distribution has to be processed. Specifically, we considered an exponential weighted moving average (EWMA) based algorithm for the estimation of CSs in non-stationary conditions \cite{Raza2013c}. To assess the performance of the proposed algorithm, this study extensively compared the proposed method with various existing passive ensemble learning algorithms: Bagging, Boosting, and Random Subspace; and an active ensemble learning via linear discriminant analysis (LDA)-score based probabilistic classification. A series of experimental evaluations have been performed on two publicly available MI related EEG datasets.\\

The contributions of the paper are summarized as follows:
\begin{itemize}
 \item	{An active adaptive ensemble learning algorithm is proposed wherein new classifiers are added online to the ensemble based on covariate shift estimation.}
 \item  {The adaptation is performed in unsupervised mode using transduction via PW$K$NN classification.}
 \item	{The proposed system is applied to motor imagery based BCI to better characterise the non-stationary changes that occur across and within different sessions.}
\end{itemize}

The remainder of this paper proceeds as follows: Section II presents background information for CS,  NSL methods in BCI and ensemble learning methods. Section III details the proposed methodology for estimating the CSs and related adaptive ensemble algorithm. Section IV describes the proposed MI related BCI system, and gives a description of the datasets and the signal processing pipeline. Next, Section V presents the performance analysis. Finally, the results are discussed in Section VI and Section VII summarises the findings of this study.

\section{Background}
\label{sec2_back}

\begin{figure*}[t]
\centering
\includegraphics[width=11cm,height=6cm]{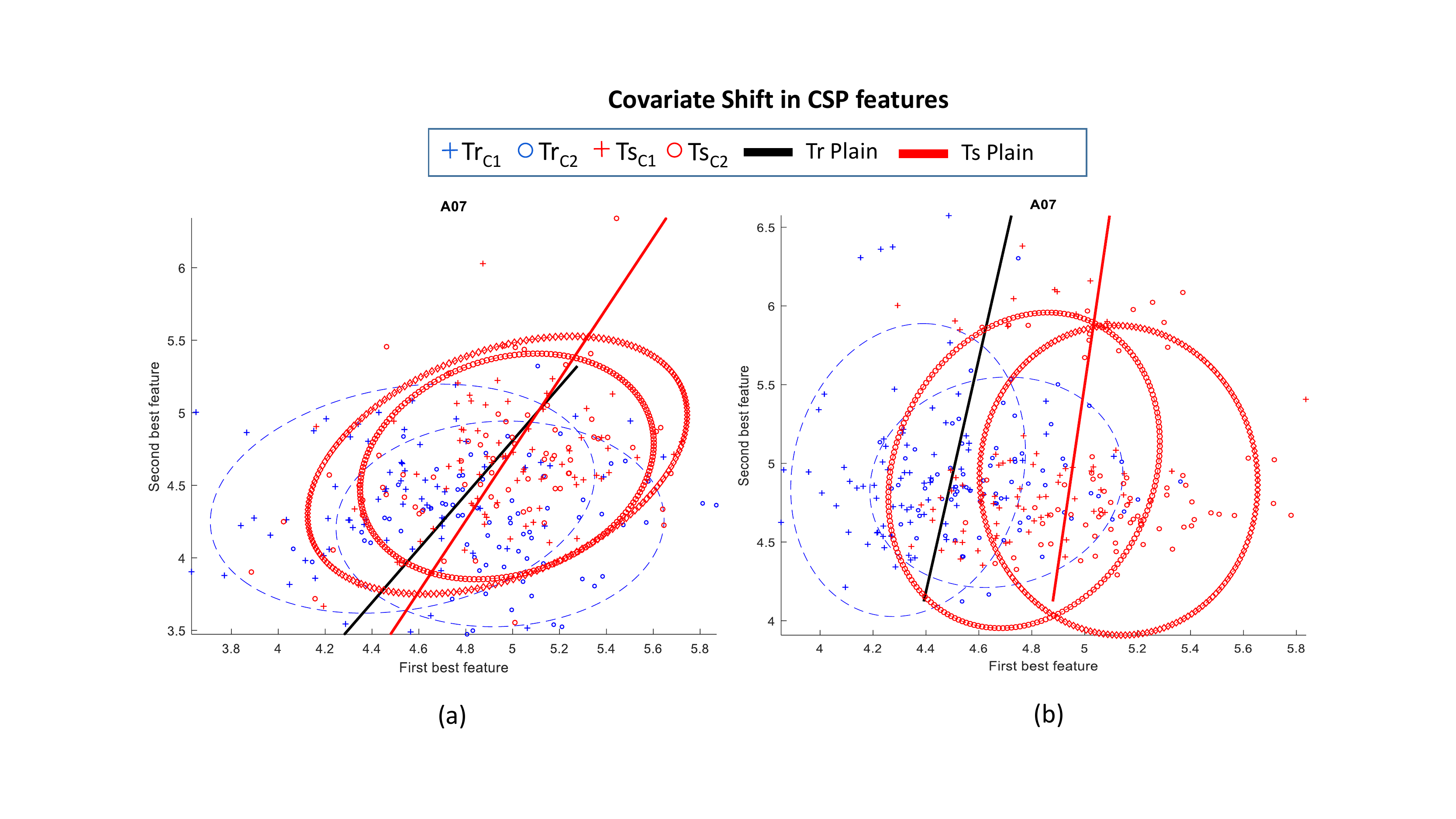}
\vspace{-1cm}
\caption{Covariate shift (CS) between the training $(Tr)$ and test $(Ts)$ distributions of subject $A07$ in dataset-2A. $(a)$ illustrates the CS in the mu $(\mu)$ band and $(b)$ shows the CS in the beta $(\beta)$ band.}
\end{figure*}

\subsection{Covariate Shift in EEG Signals}
\label{sub_sec_2.1}

In a typical BCI system, CS is a case where the input distribution of the data shifts i.e. $(P_{train}(x)\neq P_{test}(x))$, whereas the conditional probability remains the same i.e. $(P_{train} (y\vert x)=P_{test}(y\vert x)$, while transitioning from the training to testing stage. Fig. 1 illustrates the CS presence in EEG data of the subject $A07$ in dataset-2A (the description of the dataset is present in section IV). The blue solid ellipse shows the training distribution $P_{train}(x)$ and blue solid line presents the classification hyperplane for training dataset. Similarly, the red dashed ellipse shows the test distribution $P_{test}(x)$ and the red dash line presents the classification hyperplane for the test dataset. Fig.1.(a) and Fig.1.(b) provide the CSP features for $(\mu)$ band $[8-12]$ Hz and beta $(\beta)$ band $[14-30]$ Hz, respectively.  

\subsection{Non-Stationary Learning in EEG-based BCI}
\label{sub_sec_2.2}

The low classification accuracy of the existing BCI systems has been one of the main concerns in their rather low uptake among people with a severe physical disability \cite{Suk2013}. To enhance the performance of MI related BCI systems, various signal processing methods have been proposed to extract effective features in the temporal and spatial domains that can characterise the non-stationarity in EEG signals. For example, in the temporal domain, band-power and band-pass based filtering methods are commonly used \cite{Blankertz2008b}, whereas in the spatial domain, common averaging, current source density \cite{Rathee2017a}, and CSP-based features have been examined for the detection of MI related responses \cite{Ramoser2000, Raza2015b}.\\

Machine learning researchers have made efforts to devise adaptive BCI systems by incorporating NSL mechanisms into adaptation to improve the performances. Vidaurre et al. \cite{Vidaurre2006} have developed a classifier using an adaptive estimation of information matrix. Shenoy et al. \cite{Shenoy2006} have provided quantified systematic evidence of statistical differences in data recorded during multiple sessions and various adaptive schemes were evaluated to enhance the BCI performance. A CS minimization method was proposed for the non-stationary adaptation to reduce  feature set overlap and unbalance for different classes in the feature set domain \cite{Satti2010}. More interestingly, Li et al.(2010) has proposed an unsupervised CS adaptation based on a density ratio estimation technique\cite{Li2010a}. There exists a limitation that the density ratio based adaptation method requires all the testing unlabeled data before starting the testing phase to estimate the importance for the non-stationarity adaptation. This makes the approach impractical in  real-time BCI applications such as communication or rehabilitation \cite{Chowdhury2017}. To tackle these challenges, ensemble machine learning has emerged for NSL, where a set of classifiers is coupled to provide an overall decision. The generalization of an ensemble is much better than that of a single classifier \cite{Dietterich2000}, which has strong theoretical support due to the following reasons. First, in case where the training data does not provide adequate information for selecting a single optimal learner, combining classifiers in the ensemble may be a better choice. Second, the search method of best hypothesis in the source domain of a single classifier may be sub-optimal. An ensemble may compensate for such sub-optimal search process by building multiple classifiers. Third, searching true target function in the hypothesis space may not result in single optimal function, ensembles provide more acceptable approximations. In the EEG-based BCI systems, ensemble learning methods have been evaluated to improve the classification performance (e.g. bagging, boosting, and random subspace \cite{Sun2007}). Impressively, a dynamically weighted ensemble classification (DWEC) method has been proposed to handle the issue of non-stationarity adaptation \cite{Liyanage2013}. The DWEC method partitions the EEG data using clustering analysis and subsequently train multiple classifiers using the partitioned datasets. The final decision of the ensemble is then obtained by appropriately weighting the classification decisions of the individual classifiers. In a recent study, the ensemble of common spatial pattern patches has shown a potential for improving online MI related BCI system performance\cite{Sannelli2016}.\\

The above-mentioned methods were all based on the passive scheme to NSL for EEG signals. Moreover, both single classifier and classifier ensemble based approaches were developed using the passive mechanism to improve the MI detection performance. However, in passive scheme based ensemble learning, devising the right number of required classifiers to achieve an optimal performance and reducing the computational cost for adding a classifier in the ensemble during the evaluation phase are still major open challenges. Our previous study \cite{Raza2015,Raza2015_PR} demonstrated that the active scheme based learning BCI system has the potential of improving its performance. We have shown that a single active inductive classifier in single-trial EEG classification outperformed the existing passive scheme, although the developed system was only applicable for the rehabilitative BCI systems. 

\subsection{Ensemble Learning Methods in BCI Systems}
\label{sub_sec_2.3}

This study compare the proposed method with five state-of-the-art ensemble learning methods, namely Bagging, AdaBoost, TotalBoost, RUSboost, and Random Subspace. These ensemble learning methods are briefly described thereafter.

\subsubsection{Bagging}

Bagging is an ensemble machine learning meta-algorithm that involves the process of Bootstrap Aggregation ~\cite{Breiman1996}. This algorithm is a special case of the model averaging technique wherein each of the sampled datasets is used to create a different model in the ensemble and the output generated from each model is then combined by averaging (in the case of regression) or voting (in the case of classification) to create a single output. Nevertheless, bagging has the disadvantage of being ineffective in dealing with unstable nonlinear models (i.e. when a small change in the training set can cause a significant change in the model). Ensemble classification with Bagging algorithm has been applied to a P300-based BCI, and demonstrated some improvement in performance of the ensemble classifier with overlapped partitioning that requires less training data than with naive partitioning ~\cite{Onishi2014}.

\subsubsection{AdaBoost}

Boosting is a widely used approach to ensemble learning. It aims to create an accurate predictive model by combining various moderately weak classifiers. In the family of boosting methods, a powerful ensemble algorithm is Adaptive Boosting (i.e. AdaBoost) \cite{Freund1997}. It explicitly alters the distribution of training data and feeds to each classifier independently. Initially, the weights for the training samples are uniformly distributed across the training dataset. However, during the boosting procedure, the weights corresponding to the contributions of each classifier are updated in relation to the performance of each individual classifier on the partitioned training dataset. Recently, the boosting method has been employed for enhancement of MI related classification of EEG in a BCI system \cite{An2014}. It used a two-stage procedure: (i) training of weak classifiers using a deep belief network (DBN) and (ii) utilizing AdaBoost algorithm for combining several trained classifiers to form one powerful classifier. During the process of constructing DBN structure, many RBMs (Restrict Boltzmann Machine) are combined to create the ensemble. It can be less prone to the over-fitting that most learning algorithms suffer from \cite{Skurichina2002}. An improvement of $4\%$ in classification accuracy was achieved for certain cases by using the DBN based AdaBoost method. Nevertheless, AdaBoost has several shortcomings, such as its sensitivity to noisy data and outliers. 

\subsubsection{TotalBoost}

TotalBoost generates ensemble with innumerable learners having weighting factor that are orders of magnitude smaller than those of other learners \cite{Ratsch2008}. It manages the members of the ensemble by removing the least important member and then reshuffle the ensemble reordering from largest to smallest. In particular, the number of learners is self-adjusted. 

\subsubsection{RUSBoost}

RUSBoost is a boosting algorithm based on the AdaBoost.M2 algorithm \cite{Seiffert2010}. This method combines random under-sampling (RUS) and boosting for improving classification performance. It is one of the most popular and effective techniques for learning non-stationary data. Recently, its application to automatic sleep staging from EEG signals using wavelet transform and spectral features has been proposed wherein the RUSBoost method has outperformed bagging and other boosting methods \cite{Hassan2016}. However, bagging and boosting methods both have the disadvantage of being sensitive to noisy data and non-stationary environments.

\subsubsection{Random Subspace Method}

The Random Subspace Method (RSM) is an ensemble machine learning technique that involves the modification of training data in the feature space \cite{Hosseini2016,Skurichina2002}. RSM is beneficial for data with many redundant features wherein better classifiers can be obtained in random subspaces than in the original feature space. Recently, RSM method has been used in real-time epileptic seizure detection from EEG signals \cite{Hosseini2016}, where the feature space has been divided into random subspaces and the results of different classifiers are combined by majority voting to find the final output. However, RSM has a drawback as the features selection does not guarantee that the selected features have the necessary discriminant information. In this way, poor classifiers are obtained that may deteriorate the performance of ensemble learning.\\

The above-mentioned ensemble methods for the EEG classification somehow manage non-stationarity in EEG signals, but they are suitable only for passive scheme based settings wherein the ensemble has to be updated continuously over time.


\section{The Proposed Methodology}
\label{sec3_prop_method}

\subsection{Problem Formulation}
\label{sub_sec_3.1}

Given a set of training samples $X^{Train}=\big\{x_i^{train},y_i^{train} \big\}$, where $i \in \big\{1...n \big\}$ is the number of training samples, $x_i^{train}\in\mathbb{R}^D$ ($D$ denotes the input dimensionality) is a set of training input features drawn from a probability distribution with density $P_{train}(x)$, and $y_i^{train}\in \big\{C_1,C_2\big\}$ is a set of training labels, where $y_i=C_1$, if $x_i$ belongs to class  $\omega_1$, and $y_i=C_2$, if $x_i$ belongs to class $\omega_2$. We assumed that the input training data distribution remains stationary during the training phase. In addition to the labeled training samples, let's assume unlabeled test input observations $X^{Test}=\big\{x_i^{test}\big\}$, where $i \in \big\{1...m \big\}$ is the number of testing observations, $x_i^{test}\in\mathbb{R}^D$ is a set of test input features, drawn independently from a probability distribution with density $P_{test}(x)$. Note that we consider the CS presence in the data and thus, the input distributions may be different during the training and testing phases (i.e. $P_{train}(x) \neq P_{test}(x)$). 

\subsection{Covariate Shift Estimation}
\label{sub_sec_3.2}

The CS estimation (CSE) is an unsupervised method for identifying non-stationary changes in the unlabeled testing data $(X^{Test})$ during the evaluation phase \cite{Raza2015_PR}. The pseudo code is presented in Algorithm 1. The parameters for the CSE are predetermined during the training phase. The CSE algorithm works in two stages. The first stage is a retrospective stage wherein an $(EWMA)$ model is used for the identification of the non-stationarity changes in the streaming data. The EWMA is a type of infinite impulse response filter that applies weighting factors which decrease exponentially. The weight of each older observation decreases exponentially, however, never reaching zero values. The weighting factor is one of the strengths of the EWMA model. The EWMA control chart overtakes other control charts because it pools together the present and the past data in such a way that even small shifts in the time-series can be identified more easily and quickly. Furthermore, the incoming observations are continuously examined to provide $1$-step-ahead predictions and consequently, $1$-step-ahead prediction errors are generated. Next, if the estimated error fell outside the control limits ($L$), the point is assessed to be a CS point. The EWMA model presented in Eq. (1), is used to provide a $1$-step-ahead prediction for each input feature vector of the EEG signals.

\begin{equation} 
z_{(i)}=\lambda x_{(i)}+(1-\lambda)z_{(i-1)}
\end{equation}

where $\lambda$ is a smoothing constant to be selected based on minimizing $1$-step-ahead-prediction error on the training dataset $(X^{Train})$. The selection of the value of $\lambda$ is a key issue in the CSE procedure. Specifically for the auto-correlated time series data, it was suggested to select a value of $\lambda$ that minimized the sum of the squares of the 1-step ahead prediction ($1$-SAP) errors \cite{Montgomery1991}. However, we incorporated data-driven approach and thus, the optimum value of $\lambda$ was obtained by testing different values of $\lambda$ in the range of $[0 1]$ with a step of 0.01 on the training dataset. The second stage was a validation stage wherein the CS warning issued at first stage was further validated. A multivariate two-sample Hotelling's T-Square statistical hypothesis test was used to compare two distinct samples of equal number of observations generated before at the CS warning time point. If the test rejected the null hypothesis, the existence of CS was confirmed via this stage, otherwise, it was considered as a false alarm \cite{Raza2016_SC}. 

 
\begin{algorithm}
 \caption{Covariate Shift Estimation (CSE) \cite{Raza2015_PR}}
\begin{algorithmic}[1] \small
\renewcommand{\algorithmicrequire}{\textbf{Input}}
\renewcommand{\algorithmicensure}{\textbf{Output}}
\REQUIRE $: X^{Train}$, $X^{Test}$ \\
\ENSURE  $: p-value $\\ 
\textit{\textbf{Set the following parameters on training dataset}}:
  \STATE Set the following parameters on training dataset :- $z_{0}$: arithmetic mean of training input, $\lambda$: smoothing constant , $\sigma_{err^{2}_{0}}$: standard deviation of the 1-step-ahead-predicted error using unlabeled training data, and $PW$: transformation matrix from principal component analysis (PCA). For more details (see ~\cite{Raza2015_PR})\\
\textit{\textbf{Start testing phase}}:
	\FOR {$i = 1$ to $m$ in $X^{Test}$}  
	\STATE $x_{(i)}$=$PW \times x_{(i)}$ \# Get the $1^{st}$ component
  \STATE $z_{(i)}=\lambda.x_{(i)}+(1-\lambda).z_{(i-1)}$ \# Compute the $z$-statistics 
	\STATE $err_{(i)}=x_{(i)}+z_{(i-1)}$ \#  Compute 1-SAP error 
	\STATE $\widehat{\sigma}_{err^2_{(i)}}=\vartheta.err_{(i)}+(1-\vartheta).\widehat{\sigma}_{err^2_{(i-1)}}$ \# Compute smoothed variance 
	\STATE $UCL_{(i)}=z_{(i-1)} + L.\sqrt{\widehat{\sigma}_{err^2_{(i-1)}}}$
	\STATE $LCL_{(i)}=z_{(i-1)} - L.\sqrt{\widehat{\sigma}_{err^2_{(i-1)}}}$
	\IF {$LCL_{(i)}\leq x_{(i)}\leq UCL_{(i)}$}
			\STATE no shift			
			\ELSE 
			\STATE Issue CS warning and go to stage-II (i.e. CSV)
			\STATE Stage-II: execute Hotelling T-squared test on the current feature vector and average feature vector of $X^{Train}$ to get $p$-value
	\ENDIF
  \ENDFOR
 \RETURN $p$-value 
 \end{algorithmic} 
 \end{algorithm}  
 

\begin{algorithm}[p]
 \caption{CSE-UAEL}
 \begin{algorithmic}[1] \small
 \renewcommand{\algorithmicrequire}{\textbf{Input}}
 \renewcommand{\algorithmicensure}{\textbf{Output}}
 \REQUIRE $: X^{Train}=\big\{x_i^{train},y_i^{train} \big\}$, where $i \in \big\{1...n \big\}$ \\
          $: X^{Test}=\big\{x_i^{test} \big\}$ where $i \in \big\{1...m \big\}$
 \ENSURE  $: Y^{Test}$ $\mbox{and}$  $Mean Square Error$ \\ 
  \textit{\textbf{TRAINING}}:
  \STATE $E \leftarrow \emptyset$
	\STATE $f_1\leftarrow Train(X^{Train})$
	\STATE $E\leftarrow E\cup f_1$\\
  \textit{\textbf{TEST}}:
  \STATE Start evaluation using testing dataset $X^{Test}$
  \STATE Set $i,k=1$, where $k$ is the cardinality of ensemble $E$
  \STATE $\hat{y}_i^k=E(x_i)$
  \FOR {$i = 2$ to $m$} 
  \IF {($CSE(X_{i}^{Test}$)$<0.05$) $\#$ See Algorithm 1}
  \STATE $k=k+1$
	\STATE $X^{New}\leftarrow \emptyset$
	\STATE $X^{Temp}=\big\{ \big(x_v^{test}\big) \big\}_{v=1:i}$
  \FOR {$j = 1$ to $i$}
			\STATE [$CR$] $\leftarrow$ PW$K$NN($X^{Temp}_{j}$, $X^{Train}$, $K$, $\kappa$) \# See Algorithm 3
			\IF {($CR$ $>$ $\Gamma$)}
			\STATE Add $X^{Temp}_{j}$ and Predicted label to $X^{New}$
			\ELSE 
			\STATE Reject trial $X^{Temp}_{j}$
			\ENDIF
  \ENDFOR	
	\STATE $X^{Train}=(X^{Train} \cup X^{New})$
	\STATE $f_{k}\leftarrow Train(X^{Train})$
	\STATE $E\leftarrow E\cup f_k$
  \ENDIF
	\STATE $\hat{y}_i^k=E(x_i)$
	\STATE $\hat{y}_i^{test}=\sum_{k=1}^{end}\hat{y}_i^{k}$
  \ENDFOR
 \RETURN $Y^{Test}$ 
 \end{algorithmic} 
 \end{algorithm}

\begin{algorithm}[p]
 \caption{PWKNN}
 \begin{algorithmic}[1] \small
 \renewcommand{\algorithmicrequire}{\textbf{Input}}
 \renewcommand{\algorithmicensure}{\textbf{Output}}
 \REQUIRE $: x_{p}, X^{Train}, K, \kappa$ \\
 \ENSURE  $: CR$ \\
  \STATE Select $K$-nearest neighbour from $X^{Train}$ into $X^{q}=\big\{x_z,y_z\big\}$, where $z \in \big\{1...K \big\}$
  \STATE $CR_{\omega_{(1)}}:: P(\omega_{(1)}|x_p)=\frac{\sum_{j=1}^{K} \kappa(x_p,x_j)*(y_j==\omega_{(1)})}{\sum_{j=1}^{K} \kappa(i)}$ \# $\kappa$ was a function, see Eq. 6.
	\STATE $CR_{\omega_{(2)}}:: P(\omega_{(2)}|x_p)=1 - CR_{\omega_{(2)}}$
 \RETURN $CR=max(CR_{\omega_{(1)}}$, $CR_{\omega_{(2)}})$ 
 \end{algorithmic} 
 \end{algorithm}

\subsection{CSE-based unsupervised adaptive ensemble learning (CSE-UAEL)}

The CSE-UAEL algorithm combined the aforementioned CSE procedure and an unsupervised adaptation method using a combination of transductive-inductive approach. The pseudo code of CSE-UAEL is described in the Algorithm 2. The core idea of the proposed algorithm is to adapt to the non-stationary changes by using both the information from the training dataset and the new knowledge obtained in unsupervised mode from the testing phase. 

The transductive method is used to add new knowledge in the existing training dataset $(X^{Train})$ during the testing phase, wherein a probabilistic weighted $K$ nearest neighbour (PW$K$NN) method (i.e. instance based learning) \cite{Kasabov2003} is implemented and the ensemble of inductive classifiers $(E)$ is used for predicting the BCI outputs. Each time a CS is identified using the CSE procedure (Algorithm 2, step 8), a new classifier is added to the ensemble based on the updated training dataset (Algorithm 2, step 22). The training dataset is updated at step 20 (Algorithm 2) without considering the actual labels of the testing data and to adapt to the evolution of CS over time in the feature set of the testing phase. The output from the PW$K$NN method (i.e. $CR$ at step 13) is used to determine whether a trial and its corresponding estimated label can be added to the training dataset and subsequently, the learning model is updated. If the $CR$ is greater than the previously estimated threshold $\Gamma$ (cf. 4.3) then only the features of the current trial and estimated label are added to the $X^{New}$ at step 15 and the end of the for loop the new classifier is trained on the updated $X^{Train}$ (step 21). This procedure is repeated at each identified CS point and trials are added to the initial training dataset along with addition of a new and updated classifier to the current ensemble at step 22. Transductive learning via PW$K$NN combines induction and deduction in a single step and is related to the field of semi-supervised learning (SSL), which used both labeled and unlabeled data during learning process \cite{Raza2014c,Raza2016_IJCNN}. Thus, by eliminating the need to construct a global model, transductive method offerd viable solution to achieve a higher accuracy. However, in order to make use of unlabeled data, it is necessary to assume some structure to its underlying distribution. Additionally, it is essential that the SSL approach must satisfy at least one of the following assumptions such as smoothness, cluster, or manifold assumption \cite{Zhu2005}. The proposed algorithm makes use of the smoothness assumption (i.e. the points which are close to each other are more likely to share the same label) to implement the PW$K$NN algorithm. The pseudo code of the PW$K$NN algorithm is given in Algorithm 3. 

\paragraph{Probabilistic Weighted $K$ Nearest Neighbor}

A $K$-nearest-neighbors ($K$NN) (i.e. a transductive learning method) based non-parametric method is used to assess current test observations. The $K$NN algorithm belonged to a family of instance-based learning methods. In this case, a small sphere centered at the point $x$ is used, where the data density $P(x)$ should be estimated. The radius of the sphere is allowed to grow until it contained $K$ data points and the estimate of the density is given by:

\begin{equation}
P(x) = \frac{K}{N' \cdot V}                                                
\end{equation}																		

\noindent where the value of \textit{V} is set to equal to the volume of the sphere, and $N'$ is the total number of data points. The parameter $K$ governed the degree of smoothing. The technique of $K$NN density estimation can be extended to the classification task in which the $K$NN density estimation is obtained for each class and the Bayes' theorem is used to perform a classification task. Now, assuming that a dataset comprised of $N'_{\omega_i}$ points in the class $\omega_i$ within the set of classes $\omega$, where $i \in \{1,2\}$, so that $ N'=\sum_{i} N'_{\omega_i}$. To classify a new point $x$, a sphere centered on $x$ containing precisely $K$ points is used irrespective of their classes. Now suppose this sphere has the volume $V$ and contains $K_{\omega_i}$ from class $\omega_i$. Then, an estimate of the density associated with each class or likelihood can be obtained by:

\begin{equation}
P(x|\omega_i) = \frac{K_{\omega_i}}{N'_{\omega_i} \cdot V}
\end{equation}

Similarly, the unconditional density is given by $P(x)=K/(N'\cdot V)$, whereas the class prior probability is given by:

\begin{equation}
P(\omega_i) = \frac{N'_{\omega_i}}{N'}
\end{equation}						
																																			
\noindent Now, using the Bayes' theorem, we can obtain the posterior probability of the class membership by using following equation:

\begin{equation}
P(\omega_i|x) = \frac{P(x|\omega_i) P(\omega_i)}{P(x)} = \frac{K_{\omega_i}}{K}   
\end{equation}

\noindent To minimize the probability of misclassification, one needed to assign the test point $x$ to the class $\omega_i$ with the largest posterior probability, i.e. corresponding to the largest value of $K_{\omega_i}/K$. Thus, to classify a new point, one needed to identify the $K$-nearest points from the training dataset and then assign the new point to the set having the largest number of representatives. This posterior probability is known as the Bayesian belief or confidence ratio ($CR$). However, the overall estimate obtained by the $K$NN method may not be satisfactory, because the resulting density is not a true probability density since its integral over all the samples space diverges \cite{Bishop2006}. Another drawback is that it considers only the $K$ points to build the density and thus, all neighbors have equal weights. An extension to the above $K$NN method is to assign a weight to each sample that depends on its distance to $x$. Thus, a radial basis function (RBF) kernel ($\kappa$) can be used to obtain the weights, which assigns higher weights to the nearest points than furthest points (see Eq. 6). 

\begin{equation}
\kappa(x_p,x_q)=exp( -\frac{{(||x_p-x_q||)}^2}{2\sigma^2} )\\
\end{equation}

\noindent where ${{(||x_p-x_q||)}^2}$ is the squared Euclidean distance from the data point $x_p$ to the data point $x_q$ and $\sigma$ is a free parameter. For binary detection, the confidence ratio of $CR_{\omega_i}$ of the class $\omega_i$, for a data point $x_p$, is defined by:

\begin{equation}
CR_{\omega_1} = \frac{\sum\limits_{q=1}^{K} \kappa{(x_p,x_q)} \cdot (y_q == \omega_1)}{\sum\limits_{q=1}^{K} \kappa{(x_p,x_q)}}\\             
\end{equation}

\begin{equation}
CR_{\omega_2} = 1-CR_{\omega_1}
\end{equation}

\noindent where $1 \leq q \leq k$, corresponds to the $q^{th}$ nearest neighbor of $x_p$.
The outputs of PW$K$NN include the overall confidence of the decision given by:

\begin{equation}
CR = max( CR_{\omega_1},CR_{\omega_2} )
\end{equation}

\noindent and the output class $\widehat{y}$ is equals to 1 if $x_p$ is assigned to $\omega_1$ otherwise equals to 0.

\subsection{Complexity Analysis}

The core idea behind the proposed technique is to take advantage of an active scheme based NSL for initiating unsupervised adaptation by adding new classifiers to the ensemble each time a CS is identified. The choice of the classifier to be used may depend on its complexity. By considering $m$ labeled examples and $n$ examples to test, the PW$K$NN method requires a linear time (i.e. $\mathcal{O}(nmD)$) to predict the labels during testing phase as it belongs to the family of an instance based learning, whereas in other approaches such as LDA, a quadratic time is required to predict the score (i.e. $\mathcal{O}({mD}^2)$) for training the classifier, if ($m>D$), where $D$ is the dimensionality \cite{Cai2008}. For the test, LDA requires a linear time (i.e. $\mathcal{O}(nD)$). Therefore, depending on the number of trials to test after training, PW$K$NN is less computationally expensive than LDA if $n<mD/(m-1)$.



\section{Application to Motor-Imagery related BCI System}
\label{sec4_BCI_Application}

\begin{figure*}[t]
\centering
\begin{tabular}{@{}c@{}}
{\includegraphics[width=17cm,height=7.5cm]{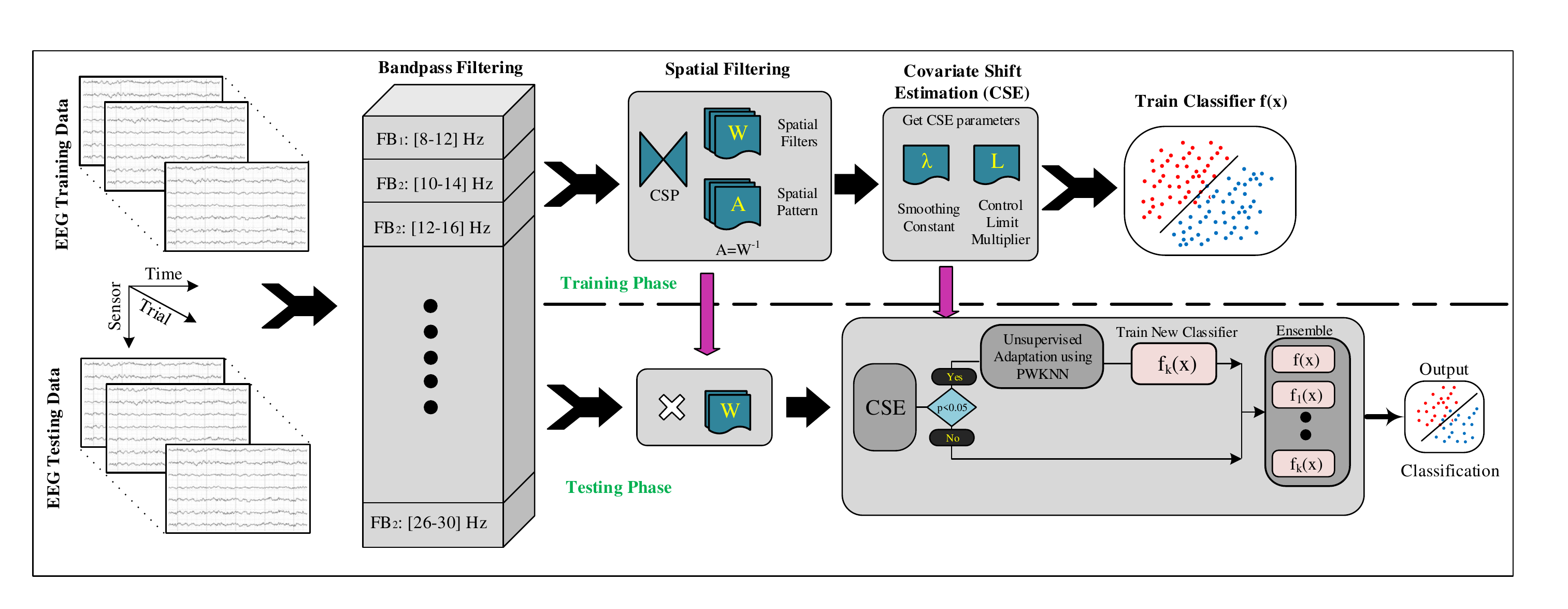}}
\end{tabular}
\vspace{-.8em}
\caption{Block diagram of the signal processing and machine learning pipeline implemented in the study. The system consists of two phases. During the training phase, the features were extracted in the temporal and spatial domains from the raw EEG signals, followed by the estimation of covariate shift parameter (i.e. $\lambda$ and L, smoothing constant and control limit multiplier, respectively) and a classifier is trained on the labeled examples (i.e. $X^{Train}$). In the evaluation phase, a similar signal processing method is applied initially and CSP features were monitored by the CSE and adaptation block. In the CSA block, the CSE procedure identifies the CSs and initiates adaptation by adding the $k^{th}$ classifier $f_k$ to the ensemble $E$, where $k$ counts the number of identified CSs during the evaluation phase. Finally, the $k$ classifier outputs from $E$ are combined to predict the class label.}
\label{fig:blockdiag}
\end{figure*}

\subsection{MI related EEG Datasets}

To assess the performance of the proposed CSE-UAEL algorithm, a series of experimental evaluations are performed on the following publicly available MI related EEG datasets. 
\subsubsection{BCI Competition IV dataset-2A}

The BCI Competition-IV dataset-2A \cite{Tangermann2012} comprising of EEG signals was acquired from nine healthy participants , namely $[A01-A09]$. The data were recorded during two sessions on separate days for each subject using a cue-based paradigm. Each data acquisition session consisted of 6 runs where each run comprised of 48 trials (12 trials for each class). Thus, the complete study involved 576 trials from both sessions of the dataset. The total trial length is 7.5~s with variable inter-trial durations. The data were acquired from 25 channels (22 EEG channels along with three monopolar EOG channels) with a sampling frequency of 250~Hz and bandpass filtered between 0.5 Hz to 100~Hz (notch filter at 50~Hz). Reference and ground were placed at the left and right mastoid, respectively. Among the 22 EEG channels, 10 channels, responsible for capturing most of the MI related activations, were selected for this study (i.e. channels: $C3$, $FC3$, $CP3$, $C5$, $C1$, $C4$, $FC4$, $CP4$, $C2$, and $C6$). The dataset consisted of four different MI tasks: left hand (class $1$), right hand (class $2$), both feet (class $3$), and tongue (class $4$). Only the classes corresponding to the left hand and right hand were considered in the present study. The MI data from the session-I was used for training phase and the MI data from the session-II was used for evaluation phase. 

\subsubsection{BCI Competition IV dataset-2B}

BCI competition $2008$-Graz dataset 2B \cite{Tangermann2012} comprising of EEG data of nine subjects, namely $[B01-B09]$ was acquired over three channels (i.e. $C3$, $Cz$, and $C4$) with a sampling frequency of $250$ Hz. EEG signals were recorded in monopolar montage with the left mastoid serving as reference and the right mastoid as ground. For each subject, data corresponding to five sessions was collected, with the trial length of 8~s. The MI data using the $3$ channels from session-$I$, $II$, and $III$ were used to train the classifiers and the data from sessions $IV$ and $V$ were merged and used for evaluation phase.

\subsection{Signal Processing and Feature Extraction}

Fig. 2 depicted the complete signal processing pipeline proposed in this study for CS estimation and adaptation of MI related EEG patterns. The following steps were executed for task detection: raw EEG signal acquisition, signal processing (i.e.  temporal filtering), feature extraction (i.e. spatial filtering), estimation of CSs, adaptation of the ensemble, and finally classification.\\

\paragraph{Temporal Filtering}
In the signal processing and feature extraction stage, a set of band-pass filters was used to decompose the EEG signals into different frequency bands (FBs) by employing an $8^{th}$ order, zero-phase forward and reverse band-pass Butterworth filter. A combination total of $10$ band-pass filters (i.e. filter bank) with overlapping bandwidths, including $[8-12]$, $[10- 14]$, $[12-16]$, $[14-18]$, $[16-20]$, $[18-22]$, $[20-24]$, $[22-26]$, $[24- 28]$, and $[26-30]$ Hz was used to process the data. 

\paragraph{Spatial Filtering}
In MI-related BCI systems, both physical and imaginary movements performed by subjects cause a growth of bounded neural rhythmic activity known as event related synchronization/desynchronization (ERD/ERS). Spatial filtering was performed using CSP algorithm to maximize the divergence of band-pass filtered signals under one class and minimize the divergence for the other class. The CSP algorithm has been widely implemented for estimation of spatial patterns related to ERD/ERS \cite{Liyanage2013}. In summary, the spatially filtered signal $Z$ of a single trial EEG is given as: 

\begin{equation}
\textbf{$Z$} = \textbf{$WE'$}
\end{equation}

\noindent where $E'$ is an $C\times T$ matrix representing the raw EEG of single trial, $C$ is number of EEG channels and $T$ is the number of samples for trial. In eq.( 11), $W$ is a projection matrix, where rows of $W$ were spatial filters and columns of $W^{-1}$ were the common spatial patterns. The spatial filtered signal $Z$ given in the above equations maximizes the differences in the variance of the two classes of EEG measurements. Next to CSP filtering, the discriminating features were extracted using a moving window of $3$ s starting from the cue onsets so as to continue our further analysis on the MI-related features only. However, the variances of only a small number $h$ of the spatial filtered signal were generally used as features for classification.The first $h$ and last $h$ rows of $Z$ i.e. $Z_p$, $p\in\{1\ldots2h\}$ from the feature vector $X_p$ given as input to the classifier (i.e. extreme left and right components of the CSP filter). Finally, the obtained features from all FBs were merged to create the set of input features for the classification. 

\begin{equation}
X_p=log\left( \frac{var(Z_p)}{\sum_{i=1}^{2h} var(Z_p)}\right)
\end{equation}

\subsection{Feature Selection and Parameter Selection}

The existing training dataset was further partitioned into 70$\%$ for training data subsets and 30$\%$ for validation data subsets, where validation samples were used to estimate the parameters of the proposed method. In order to estimate the CSs with the obtained multivariate inputs features, the PCA was used to reduce the dimensionality of the feature set \cite{Kolter2003}. PCA provided fewer components, containing most of the variability in the data. Next, the CSE method was applied to the PCA output features for identifying CS points at the first stage of the CSE procedure. A moving window of $3$ s of CSP features after the cue onset in the current trial was extracted to use as a first sample and a window of averaged CSP features from training data was used as the second sample in the multivariate two-sample Hotelling's T-Square statistical hypothesis test. In the CSE-UAEL algorithm, the subject specific parameters such as $K$ and $\mathbb{T}$ were selected on validation dataset using grid search method to maximize the accuracy. 

\subsection{Evaluation of Performance}

The performances of CSE-UAEL algorithm with both single and ensemble of classifiers were evaluated with the passive and active schemes to NSL in unsupervised adaptation scheme. With single classifier and ensemble based methods, both active and passive schemes were employed with the unsupervised adaptation. In the passive scheme, adaptation was performed after every $10$ trials, whereas in the active scheme, the adaptation was achieved after each CS confirmation. In both passive and active schemes, unsupervised adaptation was performed using three possible combinations of classifiers. First, combination-1 (C-1) used PW$K$NN method in both stages i.e., for enriching the training dataset and classification during testing phase. Second, combination-2 (C-2) used inductive LDA classifier for the BCI output, where the posterior probability of two classes obtained using LDA was used to determine if the trial needed to be added to enrich the training data at each CSs identification in active scheme. In C-2, the ensemble of LDA classifiers gave the combined decision using weighted majority voting scheme. Finally, combination-3 (C-3) used transductive method, where the CR of two classes against the $\mathbb{T}$, obtained using PW$K$NN method, was used to determine if the trial needed to be added to enrich the training dataset and the ensemble of LDA classifiers gave the combined decision using weighted majority voting scheme. Thus, C-3 was a combination of transductive-inductive learning. Likewise, ensemble method was implemented for both the passive and active schemes, where the ensemble was updated with a new classifier after every $10$ trials (in case of passive scheme) or at the instances of identifying CS (in case of active scheme). The parameter estimation remained same for all the combinations. Moreover, the results obtained by the proposed method for the dataset-2A was compared with the state-of-the-art methods for non-stationary adaptation in EEG such as common spatial pattern (CSP) \cite{Ramoser2000}, common spatial spectral pattern (CSSP) \cite{Lemm2005}, filter bank CSP (FBCSP) \cite{Ang2008}, optimal spatio-spectral filter network with FBCSP (OSSFN-FBCSP) \cite{Zhang2011}, and  recurrent quantum neural network (RQNN) \cite{Gandhi2014}.\\

\begin{figure}[h]
\centering
\begin{tabular}{@{}c@{}}
{\includegraphics[width=6cm,height=5cm]{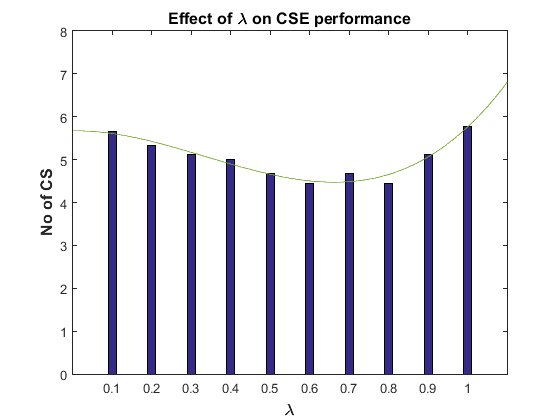}} 
\end{tabular}
\caption{The plot showed the effect of lambda ($\lambda$) on the performance of CSE at CSV stage. The average CSs identified for all the nine subjects were presented for dataset-2A.}
\label{fig:lambdaplot}
\end{figure}

\begin{table}[h]
\renewcommand{\arraystretch}{1.3}
\caption{Results for CSE procedure in dataset-2A AND dataset-2B on BCI-Competition-IV.}
\label{table:p1}
\centering
\begin{tabular}{cccc|cccc}	\toprule
\multicolumn{4}{c}{CSE for 2A} & \multicolumn{4}{c}{CSE for 2B} \\ \midrule
								    	{\textbf{Subject}} & {\textbf{$\lambda$}} & {\textbf{CSW}} &	{\textbf{CSV}} &
										  {\textbf{Subject}} & {\textbf{$\lambda$}} & {\textbf{CSW}} &	{\textbf{CSV}}\\ \midrule																				
\textbf{A01}&	0.50&	12&	6&	\textbf{B01}&	0.28&	14&	10\\
\textbf{A02}&	0.55&	15&	8&	\textbf{B02}&	0.17&	18&	13\\
\textbf{A03}&	0.60&	7&	6&	\textbf{B03}&	0.60&	19&	12\\
\textbf{A04}&	0.61&	10&	3&	\textbf{B04}&	0.20&	11&	6\\
\textbf{A05}&	0.72&	13&	8&	\textbf{B05}&	0.10&	12&	8\\
\textbf{A06}&	0.54&	12&	6&	\textbf{B06}&	0.33&	22&	12\\
\textbf{A07}&	0.57&	11&	4&	\textbf{B07}&	0.30&	17&	11\\
\textbf{A08}&	0.50&	11&	5&	\textbf{B08}&	0.21&	27&	14\\
\textbf{A09}&	0.70&	6&	4&	\textbf{B09}&	0.45&	18&	7\\ \midrule
\textbf{Mean}& 0.58& 10.77&	5.55& \textbf{Mean}& 0.29& 17.55&	10.33\\ \toprule
\end{tabular}
\end{table}

The performance analysis was based on classification accuracies (in $\%$) for binary classification tasks (i.e. Left vs Right Hand MI). Moreover, for the CSE, the number of classifiers added to the ensemble for each subject at stage-I and stage-II has been measured along with the values of $\lambda$. A two-sided Wilcoxon signed rank test was used to assess the statistical significance of the improvement at a confidence level of $0.05$ in all the pairwise comparisons. The system was implemented in MATLAB V8.1 (The Mathworks, Natick, MA) and tested on an Intel Core $i7-4790$ with $16$ GB of memory.\\

\section{Experimental Results}
\label{sec5_results}

\subsection{CSE Evaluation on Datasets-2A and -2B}

To evaluate the efficiency of the CSE procedure, a sequence of exploratory assessments was conducted on dataset-2A and -2B. Table I provides the estimated values of $\lambda$ and the corresponding number of CSs identified for both datasets during stage-I (i.e. CSW) and stage-II (i.e. CSV). The values of $\lambda$ were obtained by minimizing the sum of squares of 1-SAP errors. Moreover, Fig. 3 shows the performance of CSE at different values of $\lambda$, where the average CSs identified for all the nine subjects are presented for dataset-2A. The average number of identified CSs is 5.2, where the average of selected $\lambda$ values is $0.60$. In dataset-2A, the maximum and minimum number of identified CSs are obtained with subject $A02$ (i.e. $15$), and subject $A09$ (i.e. $6$), respectively. After the validation procedure at stage-II (i.e., CSV stage), the number of CSW for subject $A02$ decreased from $15$ to $8$, and for subject $A09$, the amount was reduced from $6$ to $4$. On an average $10.77$ CSW were received, which were further reduced to an average of $5.55$ at the CSV stage. For dataset-2B, with the combined trials from session IV and V for the evaluation phase, the maximum number of CSs were identified for subject $B08$ (i.e. $27$) and minimum for subject $B04$ (i.e. $11$). After the validation procedure at stage-II, the identified CSs for subject $B08$ were decreased from $27$ to $14$, and for subject $B04$, from $11$ to $6$. The average identified CSs (across all subjects) at stage-II for dataset-2A and -2B, have been reduced from $10.77$ to $5.55$  and $17.55$ to $10.33$, respectively as compared to stage-I. On an average $17.55$ CSW were received, which were further reduced to an average of $10.33$ at the CSV stage. It can be seen that the CSV procedure at stage-II assisted to significantly reduce the number of false CSs based on the information provided by CSW at the stage-I. In this way, the attempt of initiating adaptation by adding classifiers to the ensemble became worthless without implementing stage-II. Nevertheless, for each dataset, the number of CSV at stage-II denoted the number of classifiers added to the ensemble from the beginning to the end of the evaluation phase.

\subsection{Classification based Evaluation on Dataset-2A and -2B}

As mentioned in section 4.B,  FBCSP based features were used for various binary classifications to evaluate the performances of all the competing methods and the proposed combinations. The first analysis involved implementation of a single classifier at the evaluation stage. For dataset-2A, the classification accuracies (\%) for  C-1 (i.e. PW$K$NN-PW$K$NN), C-2 (i.e. LDA-LDA), and C-3 (i.e. PW$K$NN-LDA) were presented in Table 2 for both passive and active schemes. Similarly, for the dataset-2B, classification accuracies (\%) were provided for this analysis in Table 3. In single classifier based method, combination-3 (i.e. combination of PW$K$NN-LDA) provided higher average binary classification accuracies for both the datasets i.e 2A (cf. Table 2) and 2B (cf. Table 3) and for both passive and active schemes. In contrast, combination-1 (i.e. PW$K$NN-PW$K$NN) provided lowest average binary classification accuracies in all cases. The results clearly showed better performance of PW$K$NN-LDA combination for both datasets and schemes. \\ 

Furthermore, the second analysis involved the proposed method (i.e. CSE-UAEL) using ensemble of classifiers at the evaluation stage. The results were obtained using the CSE-UAEL algorithm in both passive and active schemes against other baseline methods (i.e. Bagging, AdaBoost, TotalBoost, RUSBoost, and RSM) are presented in Table 5 for dataset-2A and Table 6 for dataset-2B.\\ 

\begin{table*}[p]
\caption{Classification Accuracy in (\%) for dataset-2A in both passive and active schemes. C-1: a combination of PW$K$NN-PW$K$NN classifiers; C-2: a combination of inductive-inductive classifiers (i.e. LDA-LDA); and C-3: a combination of inductive-transductive classifiers (i.e. PW$K$NN-LDA).}
\label{table:p2}
\centering
\begin{tabular}{c|ccc|ccc}	\toprule
\multirow{3}{*}{\textbf{Subjects}}&	\multicolumn{6}{c}{\textbf{Single Classifier}}\\ \cmidrule{2-7}
																&	\multicolumn{3}{c|}{\textbf{Passive Scheme}}& \multicolumn{3}{c}{\textbf{Active Scheme}}\\ \cmidrule{2-7}
																&	\textbf{C-1}&	\textbf{C-2}&	\textbf{C-3}&	\textbf{C-1}&	\textbf{C-2}&	\textbf{C-3}\\ \midrule

\textbf{A01}		&58.33	&87.50	&90.28	&58.33	&91.67	&88.89  \\
\textbf{A02}		&54.17	&58.33	&64.58	&54.17	&63.19	&63.89	\\
\textbf{A03}		&54.17	&95.83	&94.44	&54.17	&91.67	&95.14	\\
\textbf{A04}		&51.39	&67.36	&69.44	&51.39	&69.44	&69.44	\\
\textbf{A05}		&66.67	&69.44	&71.53	&65.28	&70.14	&74.31	\\
\textbf{A06}		&47.22	&65.28	&66.67	&49.31	&68.06	&65.97	\\
\textbf{A07}		&53.47	&77.08	&72.92	&53.47	&72.92	&72.92	\\
\textbf{A08}		&45.83	&86.81	&91.67	&45.83	&91.67	&92.36	\\
\textbf{A09}		&43.06	&88.89	&88.19	&41.67	&88.89	&88.19	\\ \midrule
\textbf{Mean} 	&52.70	&77.39	&78.86	&52.62	&78.63	&79.01	\\
\textbf{Std}  	&7.10	  &12.93	&12.01	&6.86	  &12.01	&12.09	\\  \toprule
\end{tabular}
\end{table*}

\begin{table*}[p]
\caption{Classification Accuracy in (\%) for dataset-2B in both passive and active schemes. C-1: a combination of PWKNN-PWKNN classifiers; C-2: a combination of inductive-inductive classifiers (i.e. LDA-LDA); and C-3: a combination of inductive-transductive classifiers (i.e. PW$K$NN-LDA).}
\label{table:p3}
\centering
\begin{tabular}{c|ccc|ccc}	\toprule
\multirow{3}{*}{\textbf{Subjects}}&	\multicolumn{6}{c}{\textbf{Single Classifier}}\\ \cmidrule{2-7}
																&	\multicolumn{3}{c|}{\textbf{Passive Scheme}}& \multicolumn{3}{c}{\textbf{Active Scheme}}\\ \cmidrule{2-7}
																&	\textbf{C-1}&	\textbf{C-2}&	\textbf{C-3}&	\textbf{C-1}&	\textbf{C-2}&	\textbf{C-3}\\ \midrule
\textbf{B01}	&50.31	&70.31	&74.06	&51.25	&66.56	&75.63	\\
\textbf{B02}	&51.35	&50.31	&50.31	&52.81	&51.15	&51.15	\\
\textbf{B03}	&48.13	&46.88	&51.88	&48.13	&50.31	&51.88	\\
\textbf{B04}	&50.00	&90.00	&92.50	&49.06	&89.06	&92.50	\\
\textbf{B05}	&54.38	&80.31	&78.13	&55.94	&74.38	&72.50	\\
\textbf{B06}	&50.63	&67.50	&78.13	&50.94	&68.75	&78.75	\\
\textbf{B07}	&55.63	&68.75	&68.13	&54.06	&70.63	&68.75	\\
\textbf{B08}	&53.75	&59.69	&73.75	&53.75	&62.50	&73.75	\\
\textbf{B09}	&51.88	&66.88	&71.25	&51.88	&69.06	&71.56	\\ \midrule
\textbf{Mean}	&51.78	&66.74	&70.90	&51.98	&66.93	&70.72	\\
\textbf{Std}	&2.39		&13.49	&13.15	&2.47	  &11.79	&12.84	\\ \toprule
\end{tabular}
\end{table*}

\begin{table*}[p]
\caption{Classification Accuracy in (\%) for dataset-2A. C-1: a combination of PWKNN-PWKNN classifiers; C-2: a combination of inductive-inductive classifiers (i.e. LDA-LDA); and C-3:performance a combination of inductive-transductive classifiers (i.e. PW$K$NN-LDA).}
\label{table:p4}
\centering
\begin{tabular}{c|ccccc|ccc|ccc}	\toprule
\multirow{3}{*}{\textbf{Subjects}} & \multicolumn{5}{c}{\textbf{Baseline Methods}}& \multicolumn{6}{c}{\textbf{Proposed Methods (CSE-UAEL)}}\\  \cmidrule{2-12}
																	 &	\multicolumn{5}{c|}{\textbf{}}&	\multicolumn{3}{c}{\textbf{Passive Scheme}}& \multicolumn{3}{c}{\textbf{Active Scheme}}\\  \cmidrule{2-12}
																	 &	\textbf{BAG}&	\textbf{AB}&	\textbf{TB}&	\textbf{RUSB}&	\textbf{RSM}&	\textbf{C-1} &\textbf{C-2}&\textbf{C-3}&	\textbf{C-1} &\textbf{C-2}&\textbf{C-3}\\ \midrule
\textbf{A01}		&86.81	&71.53	&81.94	&84.72	&84.72	&58.33	&88.89	&91.67	&58.33	&87.50	&91.67\\
\textbf{A02}		&47.92	&50.69	&50.69	&52.08	&59.03	&54.17	&59.03	&63.89	&54.17	&60.42	&63.89\\
\textbf{A03}		&90.97	&71.53	&90.28	&90.28	&90.97	&54.17	&96.53	&94.44	&54.17	&95.83	&94.44\\
\textbf{A04}		&66.67	&65.28	&68.06	&67.36	&67.36	&51.39	&68.06	&70.80	&51.39	&66.67	&72.22\\
\textbf{A05}		&65.97	&70.83	&70.83	&65.97	&54.86	&65.28	&73.61	&77.78	&65.97	&72.22	&77.08\\
\textbf{A06}		&63.89	&63.19	&63.19	&64.58	&44.44	&49.31	&66.67	&73.61	&45.83	&64.58	&75.69\\
\textbf{A07}		&74.31	&75.00	&74.31	&72.92	&70.83	&53.47	&80.56	&72.92	&53.47	&74.31	&73.61\\
\textbf{A08}		&72.92	&90.97	&88.19	&90.28	&85.42	&45.83	&89.58	&93.75	&45.83	&88.89	&94.44\\
\textbf{A09}		&91.67	&84.72	&88.89	&87.50	&87.50	&41.67	&88.89	&88.89	&41.67	&89.58	&90.28\\ \midrule
\textbf{Mean}		&73.46	&71.53	&75.15	&75.08	&71.68	&52.62	&79.09	&80.86	&52.31	&77.78	&81.48\\
\textbf{Std}		&14.42	&11.76	&13.44	&13.67	&16.53	&6.86		&12.83	&11.44	&7.32		&12.87	&11.33\\ \toprule
\end{tabular}
\end{table*}

\begin{table*}[p]
\caption{Classification Accuracy in (\%) for dataset-2B. C-1: a combination of PWKNN-PWKNN classifiers; C-2: a combination of inductive-inductive classifiers (i.e. LDA-LDA); and C-3: a combination of inductive-transductive classifiers (i.e. PW$K$NN-LDA).}
\label{table:p5}
\centering
\begin{tabular}{c|ccccc|ccc|ccc}	\toprule
\multirow{3}{*}{\textbf{Subjects}} & \multicolumn{5}{c}{\textbf{Baseline Methods}}& \multicolumn{6}{c}{\textbf{Proposed Methods (CSE-UAEL)}}\\  \cmidrule{2-12}
																	 &	\multicolumn{5}{c|}{\textbf{}}&	\multicolumn{3}{c}{\textbf{Passive Scheme}}& \multicolumn{3}{c}{\textbf{Active Scheme}}\\  \cmidrule{2-12}
																	 &	\textbf{BAG}&	\textbf{AB}&	\textbf{TB}&	\textbf{RUSB}&	\textbf{RSM}&	\textbf{C-1} &\textbf{C-2}&\textbf{C-3}&	\textbf{C-1} &\textbf{C-2}&\textbf{C-3}\\ \midrule
\textbf{B01}	&69.69	&67.50	&66.25	&53.13	&51.56	&50.31	&65.31	&77.81	&51.25	&64.69	&78.13\\
\textbf{B02}	&52.60	&52.50	&55.00	&50.83	&49.79	&51.35	&50.31	&54.27	&52.81	&51.15	&54.69\\
\textbf{B03}	&50.63	&50.00	&51.56	&50.00	&50.00	&48.13	&47.50	&52.50	&48.13	&49.38	&53.13\\
\textbf{B04}	&76.25	&74.38	&81.56	&87.81	&52.19	&50.00	&89.69	&94.38	&49.06	&90.63	&94.38\\
\textbf{B05}	&67.50	&68.75	&72.81	&71.56	&53.13	&54.38	&73.44	&85.63	&55.94	&71.56	&85.31\\
\textbf{B06}	&56.88	&56.56	&59.69	&71.56	&51.88	&50.63	&69.38	&80.00	&50.94	&68.75	&80.31\\
\textbf{B07}	&58.13	&54.38	&50.00	&53.75	&50.00	&55.63	&70.00	&71.56	&54.06	&70.94	&72.81\\
\textbf{B08}	&56.88	&58.75	&59.06	&50.94	&53.13	&53.75	&60.00	&77.81	&53.75	&64.69	&78.75\\
\textbf{B09}	&55.31	&60.94	&62.81	&57.19	&49.69	&51.88	&70.31	&74.38	&51.88	&69.06	&74.38\\ \midrule
\textbf{Mean}	&60.43	&60.42	&62.08	&60.75	&51.26	&51.78	&66.22	&74.26	&51.98	&66.76	&74.65\\
\textbf{Std}	&8.66		&8.22		&10.21	&13.21	&1.42		&2.39		&12.68	&13.57	&2.47		&12.11	&13.36\\ \toprule
\end{tabular}
\end{table*}

\begin{table*}[p]
\caption{Comparison of CSE-UAEL Algorithm using p-values on dataset-2A. The p-value denotes the Wilcoxon signed-rank test:$\textsuperscript{$\ast$}$p$<0.01$, $\textsuperscript{$\star$}$p$<0.05$.}
\label{table:p6}
\centering
\begin{tabular}{c|c|cc|ccccc|ccc}	\toprule
		       &    & \multicolumn{2}{c|}{\textbf{Single Classifier}}& \multicolumn{8}{c}{\textbf{Ensemble}}\\ \cmidrule{2-12}
		       &    & Passive	  &Active    & \multicolumn{5}{c|}{\textbf{Baseline Methods}}&\multicolumn{3}{c}{\textbf{CSE-UAEL (Passive)}} \\ \cmidrule{2-12}
		       &    & C-3	  &C-3    &BAG	  &AB   	&TB   	&RUSB   &RSM    &C1   	&C2 	  &C3 \\ \cmidrule{1-12}
		       
CSE-UAEL   &C-1 &0.0039$\textsuperscript{$\ast$}$	&0.0039$\textsuperscript{$\ast$}$	&0.0156$\textsuperscript{$\star$}$	&0.0078$\textsuperscript{$\ast$}$	&0.0078$\textsuperscript{$\ast$}$	&0.0156$\textsuperscript{$\star$}$	&0.0273$\textsuperscript{$\star$}$	&1    	&0.0039$\textsuperscript{$\ast$}$	&0.0039$\textsuperscript{$\ast$}$\\

(Active)     &C-2 &0.1016	&0.1484	&0.0781	&0.0447$\textsuperscript{$\star$}$	&0.0447$\textsuperscript{$\star$}$	&0.0469$\textsuperscript{$\star$}$	&0.0078$\textsuperscript{$\ast$}$	&0.0039$\textsuperscript{$\ast$}$	&0.0781	&0.0408$\textsuperscript{$\star$}$\\

          &C-3 &0.0234$\textsuperscript{$\star$}$	&0.0234$\textsuperscript{$\star$}$	&0.0195$\textsuperscript{$\star$}$	&0.0078$\textsuperscript{$\ast$}$	&0.0078$\textsuperscript{$\ast$}$	&0.0039$\textsuperscript{$\ast$}$	&0.0039$\textsuperscript{$\ast$}$	&0.0039$\textsuperscript{$\ast$}$	&0.1562	&0.1562\\ \toprule
\end{tabular}
\end{table*}

\begin{table*}[h]
\caption{Comparison of CSE-UAEL Algorithm using p-values on dataset-2B. The p-value denotes the Wilcoxon signed-rank test:$\textsuperscript{$\ast$}$p$<0.01$, $\textsuperscript{$\star$}$p$<0.05$.}
\label{table:p7}
\centering
\begin{tabular}{c|c|cc|ccccc|ccc}	\toprule
		       &    & \multicolumn{2}{c|}{\textbf{Single Classifier}}& \multicolumn{8}{c}{\textbf{Ensemble}}\\ \cmidrule{2-12}
		       &    & Passive	  &Active    & \multicolumn{5}{c|}{\textbf{Baseline Methods}}&\multicolumn{3}{c}{\textbf{CSE-UAEL (Passive)}} \\ \cmidrule{2-12}
		       &    & C-3	  &C-3    &BAG	  &AB   	&TB   	&RUSB   &RSM    &C1   	&C2 	  &C3 \\ \cmidrule{1-12}
		       
CSE-UAEL 	&C-1 	&0.0078$\textsuperscript{$\ast$}$	&0.0078$\textsuperscript{$\ast$}$	&0.0078$\textsuperscript{$\ast$}$	&0.0078$\textsuperscript{$\ast$}$	&0.0195$\textsuperscript{$\star$}$	&0.1641	&0.4961	&0.75	&0.0195$\textsuperscript{$\star$}$	&0.0039$\textsuperscript{$\ast$}$ \\
(Active)  &C-2 	&0.0447$\textsuperscript{$\star$}$	&0.0391$\textsuperscript{$\star$}$	&0.0742	&0.0486$\textsuperscript{$\star$}$	&0.1641	&0.0781	&0.0078$\textsuperscript{$\ast$}$	&0.0078$\textsuperscript{$\ast$}$	&0.5234	&0.0039$\textsuperscript{$\ast$}$ \\
	        &C-3  &0.0039$\textsuperscript{$\ast$}$	&0.0039$\textsuperscript{$\ast$}$	&0.0039$\textsuperscript{$\ast$}$	&0.0039$\textsuperscript{$\ast$}$	&0.0078$\textsuperscript{$\ast$}$	&0.0039$\textsuperscript{$\ast$}$	&0.0039$\textsuperscript{$\ast$}$	&0.0039$\textsuperscript{$\ast$}$	&0.0039$\textsuperscript{$\ast$}$	&0.0425$\textsuperscript{$\star$}$ \\ \toprule
\end{tabular}
\end{table*}

\begin{table*}[h]
\caption{Classification Accuracy in ($\%$) Comparison with the state-of-the-art method in dataset-2A.}
\label{table:p8}
\centering
\begin{tabular}{c|c|c|c|c|cc}	\toprule
{\textbf{CSP \cite{Ramoser2000}}} & {\textbf{CCSP \cite{Lemm2005}}} & {\textbf{FBCSP \cite{Ang2008}}} &	{\textbf{OSSFN-FBCSP \cite{Zhang2011}}} &	{\textbf{RQNN \cite{Gandhi2014}}} &	{\textbf{CSE-UAEL (Active) (C-3)}}\\ \midrule
73.46&            79.78&	           76.31&	                  76.31&                   66.59& 81.48\\ \midrule
\end{tabular}
\end{table*}

The average binary classification accuracies (i.e. $mean\pm SD$) provided by unsupervised adaptation methods for dataset-2A (cf. Table 4) are:  Bagging (BAG: $73.46\pm14.42$), AdaBoost (AB:$71.53\pm11.76$), TotalBoost (TB:$75.15\pm13.44$), RUSBoost (RUSB:$75.08\pm13.67$), and RSM ($71.68\pm16.53$). For the same dataset, the average binary classification accuracies (i.e. $mean\pm SD$) provided by CSE-UAEL in passive scheme are: C-1:$52.60\pm6.86$, C-2:$79.09\pm12.83$, and C-3:$80.86\pm11.44$ and CSE-UAEL in active scheme were : C-1:$52.31\pm7.32$, C-2:$77.78\pm12.87$, and C-3:$81.48\pm11.33$. The performances of the C-3 (i.e. LDA + PW$K$NN ) were better than the existing ensemble methods and other classifier combinations for both passive and active schemes.\\

The average binary classification accuracies (i.e. $mean\pm SD$) provided by unsupervised adaptation methods for dataset-2B (cf. Table 5) were:  Bagging (BAG: $60.43\pm8.66$), AdaBoost (AB:$60.42\pm8.22$), TotalBoost (TB:$62.08\pm10.21$), RUSBoost (RUSB:$60.75\pm13.21$), and RSM ($51.26\pm1.42$). For the same dataset, the average binary classification accuracies (i.e. $mean\pm SD$) provided by CSE-UAEL in passive scheme were: C-1:$51.78\pm2.39$, C-2:$66.22\pm12.68$, and C-3:$74.26\pm13.57$ and CSE-UAEL in active scheme were : C-1:$51.98\pm2.47$, C-2:$66.76\pm12.11$, and C-3:$74.65\pm13.36$. Similar to dataset-2A, the performances of the C-3 (i.e. LDA + PW$K$NN ) were better than the existing ensemble methods and other classifier combinations for both passive and active schemes.\\

Table 6 and 7 presented the $p$-values obtained from the statistical comparison of the CSE-UAEL in active scheme with other single-classifier and ensemble of classifiers based methods for dataset-2A and $2B$, respectively. The performance of the proposed method (i.e. CSE-UAEL in C-3) was found significantly better than Bagging, AdaBoost, TotalBoost, RUSboost and RSM. The proposed method was also found significantly better than single classifier based setting for both passive and active schemes. In dataset-2A, CSE-UAEL algorithm in active mode for C-3 was not statistically significant against CSE-UAEL algorithm in passive scheme with combination C-2 and C-3. However, the same method on dataset-2B showed significantly better result (p$<$0.05). Such analysis provided strong evidence that both CSE-UAEL algorithm with combination of inductive-transductive classifiers (i.e. PW$K$NN-LDA) performed better than the other passive and active scheme. Furthermore, the performance of the proposed method was compared with other previously published state-of-the-art-methods for dataset-2A. Table 8 presents the average classification accuracies (\%) for CSP, CCSP, FBCSP, OSSSFN-FBCSP, RQNN, and CSE-UAEL (in active scheme). Evidently, CSE-UAEL outperformed all these previously proposed methods with the highest average classification accuracy of $81.48$.

\section{Discussions}


The development of efficient machine learning methods for non-stationarity of streaming data has been considered as a challenging task. To improve the performance of MI-based BCI systems, the majority of the exiting studies have focused on techniques that extract features invariant to changes of the data without the use of time specific discriminant features. Moreover, the existing non-stationarity based machine learning methods incorporated passive schemes based on the assumption of continuous existence of non-stationarity in the streaming data. 

In this study, we have shown how an active scheme based ensemble learning can be employed to address non-stationarities of EEG signals, wherein the data distributions shift between training and evaluation phases. The main idea behind the proposed system was to take advantage of an active scheme based NSL for initiating adaptation by adding new classifiers to the ensemble each time a CS was identified instead of assuming the need to update the system at regular intervals. The CSE based active scheme assists to optimize and add new classifiers to the ensemble adaptively based upon the identified changes in the input data distribution, it does not require a trial-and-error or grid search method to select a suitable number of classifiers for obtaining an enhanced classification accuracy. More importantly, the unsupervised adaption via transduction (i.e. adaption without knowing the true labels) enables this system applicable to long sessions typically considered in the practical applications of BCIs used for both communication and rehabilitation problems.\\

Indeed, the transductive learning step during the evaluation phase involved the addition of the predicted labels to the existing training dataset. This approach ensures a continuous enrichment of the existing training dataset, which can be highly crucial to a learning algorithm suffering from a high variance. The issue of a high variance was commonly found in the EEG features of poor BCI users~\cite{Raza2015b,Gao2015}. To manage the high variability issue, adding predicted labels with high confidence may improve the prediction performance as demonstrated in the study.\\

The proposed algorithm has been extensively compared with different passive scheme based ensemble learning methods: Bagging, AdaBoost, TotalBoost, RUSBoost, and RSM. The CSE-UAEL algorithm with transductive method was used to improve classification performance against single-classifier based passive and active schemes and ensemble based passive scheme. We have shown that the CSE-UAEL algorithm provided an improvement of approximately $6-10\%$ in classification accuracies compared to other ensemble based methods for dataset-2A. And the performance improvements were statistically significant in 18 out of 20 pair-wise comparisons for the CSE-AUEL algorithm in C-3 setting. It was worth noting that the proposed methodology was not limited to BCI applications as the active scheme based ensemble learning can be applied to a wide range of dynamic learning systems where the input signals evolve over time, for example, neuro-rehabilitation and communication systems. A key challenge remains the definition of a reliable function that can determine a shift detection, and classifiers that can reliably classify the training data.\\

Although the proposed method outperforms other passive schemes, there are limitations to be considered. First, the CSE procedure has been applied to the combined CSP features of multiple frequency bands, which creates a high dimensional input vector and may affect the robustness of the CSE process. This confounding factor can be handled either by using dimensionality reduction methods or by employing multiple CSE procedures at each frequency feature vector. Second, the performance of the proposed system may be adversely affected if applied to data obtained from a large number of sessions or days of recording. In this case, a recurrent concept handling method could help to dynamically manage the number of classifiers, e.g., by replacing the old classifiers with the updated classifier in the ensemble.

\section{Conclusion}

A new active scheme based non-stationarity adaptation algorithm has been proposed to effectively account for the covariate shifts influence in an EEG-based BCI system. A synergistic scheme was defined to integrate the CS estimation procedure and ensemble learning approach with transduction to determine when new classifiers should be added to the classifier ensemble. The performance of the proposed algorithm has been extensively evaluated through comparisons with state-of-the-art ensemble learning methods in both passive and active settings. The performance analysis on two BCI competition datasets has shown that the proposed method outperforms other passive methods in addressing non-stationarities of EEG signals.

\subsection*{Acknowledgment}
S.M.Z. were supported by The Farr Institute of Health Informatics Research- CIPHER (Centre for Improvement in Population Health through E-records Research) (MR /K006525/1), the National Centre for Population Health and Wellbeing Research (CA02), and Swansea University Medical School. D.R., G.P., and H.C. were supported by the Northern Ireland Functional Brain Mapping Facility project (1303/101154803), funded by InvestNI and the Ulster University.

\clearpage

\appendix

\clearpage
\section{Symbols and Notations}
\begin{table}[h]
\caption{Symbols and Notations}
\centering
\begin{tabular}{c|l}	\toprule
Symbols and Notations & Description\\ \midrule
$x$							& Input vector\\
$y$							& Output label\\
$X^{Train}$					& Training dataset including input data $x$ and output label $y$\\
$X^{Test}$					& Test dataset including input data $x$ and output label $y$\\
$X^{Temp}$					& Temporary variable to store data in testing phase \\
$n$							& Number of training samples in training data\\
$m$							& Number of training samples in testing data\\
$D$							& Input dimensionality\\
$P_{train}(x)$					& Probability distribution of input $x$\\ 
$P_{train} (y\vert x)$				& Probability of $y$ given $x$ in training data\\
$\mu$						& Mu frequency band [8-12] Hz\\
$\beta$						& Beta frequency band [14-30] Hz\\
$C_1,C_2$					& Set of labels for Class 1 and Class 2 \\
$\omega_1$ and $\omega_2$		& Class 1 and Class 2 \\
$\mathbb{R}$					& Real number\\
$\lambda$						& lambda was a smoothing constant in covariate shift estimation\\
$z$							& EWMA statistics\\
$E$							& Ensemble of classifiers\\
$f$							& Classifier\\
$K$							& $K$ for $K$ nearest neighbour\\
$k$							& Counter for the number of classifier in ensemble\\
$\kappa$						& A radial basis function (RBF) kernel \\
$p$							& $p$-value \\
$v$							& Number of samples from starting of the testing phase to the current sample\\
$\Gamma$					& Threshold \\ 
$\cup$						& Union operation \\ 
$Np$						& Total number of points \\
$V$							& Volume \\  
$E'$                    			  	& EEG signal\\
$C$                       				& Number of channels in EEG dataset\\
$T$                       				& Number of samples per trial in EEG dataset\\ 
$W$                       				& CSP projection matrix \\
$Z$                       				& spatially filtered signal \\
$\mathcal{O}$					& Big-O notation \\\toprule
\end{tabular}
\end{table}

\clearpage
\bibliographystyle{elsarticle-num}


\begin{thebibliography}{10}
\expandafter\ifx\csname url\endcsname\relax
  \def\url#1{\texttt{#1}}\fi
\expandafter\ifx\csname urlprefix\endcsname\relax\def\urlprefix{URL }\fi
\expandafter\ifx\csname href\endcsname\relax
  \def\href#1#2{#2} \def\path#1{#1}\fi

\bibitem{Alippi2008}
C.~Alippi, M.~Roveri, Just-in-time adaptive classifiers—part i: Detecting
  nonstationary changes, IEEE Transactions on Neural Networks 19~(7) (2008)
  1145--1153.

\bibitem{Ditzler2015a}
G.~Ditzler, M.~Roveri, C.~Alippi, R.~Polikar, Learning in nonstationary
  environments: A survey, IEEE Computational Intelligence Magazine 10~(4)
  (2015) 12--25.

\bibitem{Alippi2013}
C.~Alippi, G.~Boracchi, M.~Roveri, Just-in-time classifiers for recurrent
  concepts, IEEE transactions on neural networks and learning systems 24~(4)
  (2013) 620--634.

\bibitem{Alippi2008a}
C.~Alippi, M.~Roveri, Just-in-time adaptive classifiers—part ii: Designing
  the classifier, IEEE Transactions on Neural Networks 19~(12) (2008)
  2053--2064.

\bibitem{Pan2010}
S.~J. Pan, Q.~Yang, A survey on transfer learning, IEEE Transactions on
  knowledge and data engineering 22~(10) (2010) 1345--1359.

\bibitem{Sugiyama2007}
M.~Sugiyama, M.~Krauledat, K.-R. M{\~A}{\v{z}}ller, Covariate shift adaptation
  by importance weighted cross validation, Journal of Machine Learning Research
  8 (2007) 985--1005.

\bibitem{Gong2016}
M.~Gong, K.~Zhang, T.~Liu, D.~Tao, C.~Glymour, B.~Sch{\"o}lkopf, Domain
  adaptation with conditional transferable components, in: International
  Conference on Machine Learning, 2016, pp. 2839--2848.

\bibitem{Wolpaw2002}
J.~R. Wolpaw, N.~Birbaumer, D.~J. McFarland, G.~Pfurtscheller, T.~M. Vaughan,
  Brain-computer interfaces for communication and control, Clinical
  neurophysiology 113~(6) (2002) 767--791.

\bibitem{Zhou2008}
S.-M. Zhou, J.~Q. Gan, F.~Sepulveda, Classifying mental tasks based on features
  of higher-order statistics from eeg signals in brain-computer interface,
  Information Sciences 178~(6) (2008) 1629--1640.

\bibitem{Celka2005}
P.~Celka, Neuronal coordination in the brain: A signal processing perspective
  (2005).

\bibitem{Li2010a}
Y.~Li, H.~Kambara, Y.~Koike, M.~Sugiyama, Application of covariate shift
  adaptation techniques in brain-computer interfaces, IEEE Transactions on
  Biomedical Engineering 57~(6) (2010) 1318--1324.

\bibitem{Arvaneh2013a}
M.~Arvaneh, C.~Guan, K.~K. Ang, C.~Quek, Optimizing spatial filters by
  minimizing within-class dissimilarities in electroencephalogram-based
  brain-computer interface, IEEE transactions on neural networks and learning
  systems 24~(4) (2013) 610--619.

\bibitem{Raza2015_PR}
H.~Raza, G.~Prasad, Y.~Li, Ewma model based shift-detection methods for
  detecting covariate shifts in non-stationary environments, Pattern
  Recognition 48~(3) (2015) 659--669.

\bibitem{Rathee2017}
D.~Rathee, H.~Cecotti, G.~Prasad, Single-trial effective brain connectivity
  patterns enhance discriminability of mental imagery tasks, Journal of neural
  engineering 14~(5) (2017) 056005.

\bibitem{Blankertz2008b}
B.~Blankertz, R.~Tomioka, S.~Lemm, M.~Kawanabe, K.-R. Muller, Optimizing
  spatial filters for robust eeg single-trial analysis, IEEE Signal processing
  magazine 25~(1) (2008) 41--56.

\bibitem{Raza2016_SC}
H.~Raza, H.~Cecotti, Y.~Li, G.~Prasad, Adaptive learning with covariate
  shift-detection for motor imagery-based brain-computer interface, Soft
  Computing 20~(8) (2016) 3085--3096.

\bibitem{Chowdhury2018}
A.~Chowdhury, H.~Raza, Y.~K. Meena, A.~Dutta, G.~Prasad, Online covariate shift
  detection based adaptive brain-computer interface to trigger hand exoskeleton
  feedback for neuro-rehabilitation, IEEE Transactions on Cognitive and
  Developmental Systems.

\bibitem{Raza2013b}
H.~Raza, G.~Prasad, Y.~Li, Dataset shift detection in non-stationary
  environments using ewma charts, in: Systems, Man, and Cybernetics (SMC), 2013
  IEEE International Conference on, IEEE, 2013, pp. 3151--3156.

\bibitem{Raza2013c}
H.~Raza, G.~Prasad, Y.~Li, Ewma based two-stage dataset shift-detection in
  non-stationary environments, in: IFIP International Conference on Artificial
  Intelligence Applications and Innovations, Springer, 2013, pp. 625--635.

\bibitem{Raza2014}
H.~Raza, G.~Prasad, Y.~Li, Adaptive learning with covariate shift-detection for
  non-stationary environments, in: Computational Intelligence (UKCI), 2014 14th
  UK Workshop on, IEEE, 2014, pp. 1--8.

\bibitem{Raza2016PhD}
H.~Raza,
  \href{http://ethos.bl.uk/OrderDetails.do?uin=uk.bl.ethos.695308}{Adaptive
  learning for modelling non-stationarity in eeg-based brain-computer
  interfacing}, Ph.D. thesis, Ulster University (2016).
\newline\urlprefix\url{http://ethos.bl.uk/OrderDetails.do?uin=uk.bl.ethos.695308}

\bibitem{Ramoser2000}
H.~Ramoser, J.~Muller-Gerking, G.~Pfurtscheller, Optimal spatial filtering of
  single trial eeg during imagined hand movement, IEEE transactions on
  rehabilitation engineering 8~(4) (2000) 441--446.

\bibitem{Lotte2007}
F.~Lotte, M.~Congedo, A.~L{\'e}cuyer, F.~Lamarche, B.~Arnaldi, A review of
  classification algorithms for eeg-based brain-computer interfaces, Journal of
  neural engineering 4~(2) (2007) R1.

\bibitem{Shenoy2006}
P.~Shenoy, M.~Krauledat, B.~Blankertz, R.~P. Rao, K.-R. M{\"u}ller, Towards
  adaptive classification for bci, Journal of neural engineering 3~(1) (2006)
  R13.

\bibitem{Vidaurre2006}
C.~Vidaurre, A.~Schlogl, R.~Cabeza, R.~Scherer, G.~Pfurtscheller, A fully
  on-line adaptive bci, IEEE Transactions on Biomedical Engineering 53~(6)
  (2006) 1214--1219.

\bibitem{Satti2010}
A.~Satti, C.~Guan, D.~Coyle, G.~Prasad, A covariate shift minimisation method
  to alleviate non-stationarity effects for an adaptive brain-computer
  interface, in: Pattern Recognition (ICPR), 2010 20th International Conference
  on, IEEE, 2010, pp. 105--108.

\bibitem{Liyanage2013}
S.~R. Liyanage, C.~Guan, H.~Zhang, K.~K. Ang, J.~Xu, T.~H. Lee, Dynamically
  weighted ensemble classification for non-stationary eeg processing, Journal
  of neural engineering 10~(3) (2013) 036007.

\bibitem{Raza2015}
H.~Raza, H.~Cecotti, Y.~Li, G.~Prasad, Learning with covariate shift-detection
  and adaptation in non-stationary environments: Application to brain-computer
  interface, in: Neural Networks (IJCNN), 2015 International Joint Conference
  on, IEEE, 2015, pp. 1--8.

\bibitem{Suk2013}
H.-I. Suk, S.-W. Lee, A novel bayesian framework for discriminative feature
  extraction in brain-computer interfaces, IEEE Transactions on Pattern
  Analysis and Machine Intelligence 35~(2) (2013) 286--299.

\bibitem{Rathee2017a}
D.~Rathee, H.~Raza, G.~Prasad, H.~Cecotti, Current source density estimation
  enhances the performance of motor-imagery-related brain-computer interface,
  IEEE Transactions on Neural Systems and Rehabilitation Engineering 25~(12)
  (2017) 2461--2471.

\bibitem{Raza2015b}
H.~Raza, H.~Cecotti, G.~Prasad, Optimising frequency band selection with
  forward-addition and backward-elimination algorithms in eeg-based
  brain-computer interfaces, in: Neural Networks (IJCNN), 2015 International
  Joint Conference on, IEEE, 2015, pp. 1--7.

\bibitem{Chowdhury2017}
A.~Chowdhury, H.~Raza, A.~Dutta, G.~Prasad, Eeg-emg based hybrid brain computer
  interface for triggering hand exoskeleton for neuro-rehabilitation, in:
  Proceedings of the Advances in Robotics, ACM, 2017, p.~45.

\bibitem{Dietterich2000}
T.~G. Dietterich, Ensemble methods in machine learning, in: International
  workshop on multiple classifier systems, Springer, 2000, pp. 1--15.

\bibitem{Sun2007}
S.~Sun, C.~Zhang, D.~Zhang, An experimental evaluation of ensemble methods for
  eeg signal classification, Pattern Recognition Letters 28~(15) (2007)
  2157--2163.

\bibitem{Sannelli2016}
C.~Sannelli, C.~Vidaurre, K.-R. M{\"u}ller, B.~Blankertz, Ensembles of adaptive
  spatial filters increase bci performance: an online evaluation, Journal of
  neural engineering 13~(4) (2016) 046003.

\bibitem{Breiman1996}
L.~Breiman, Bagging predictors, Machine learning 24~(2) (1996) 123--140.

\bibitem{Onishi2014}
A.~Onishi, K.~Natsume, Overlapped partitioning for ensemble classifiers of
  p300-based brain-computer interfaces, PloS one 9~(4) (2014) e93045.

\bibitem{Freund1997}
Y.~Freund, R.~E. Schapire, A decision-theoretic generalization of online
  learning and an application to boosting, Journal of computer and system
  sciences 55~(1) (1997) 119--139.

\bibitem{An2014}
X.~An, D.~Kuang, X.~Guo, Y.~Zhao, L.~He, A deep learning method for
  classification of eeg data based on motor imagery, in: International
  Conference on Intelligent Computing, Springer, 2014, pp. 203--210.

\bibitem{Skurichina2002}
M.~Skurichina, R.~P. Duin, Bagging, boosting and the random subspace method for
  linear classifiers, Pattern Analysis \& Applications 5~(2) (2002) 121--135.

\bibitem{Ratsch2008}
G.~R{\"a}tsch, M.~K. Warmuth, K.~A. Glocer, Boosting algorithms for maximizing
  the soft margin, in: Advances in neural information processing systems, 2008,
  pp. 1585--1592.

\bibitem{Seiffert2010}
C.~Seiffert, T.~M. Khoshgoftaar, J.~Van~Hulse, A.~N.~R. Boost, A hybrid
  approach to alleviating class imbalance, IEEE Transactions On Systems, Man,
  And Cybernetics—Part A: Systems And Humans 40~(1).

\bibitem{Hassan2016}
A.~R. Hassan, M.~I.~H. Bhuiyan, A decision support system for automatic sleep
  staging from eeg signals using tunable q-factor wavelet transform and
  spectral features, Journal of neuroscience methods 271 (2016) 107--118.

\bibitem{Hosseini2016}
M.-P. Hosseini, A.~Hajisami, D.~Pompili, Real-time epileptic seizure detection
  from eeg signals via random subspace ensemble learning, in: Autonomic
  Computing (ICAC), 2016 IEEE International Conference on, IEEE, 2016, pp.
  209--218.

\bibitem{Montgomery1991}
D.~C. Montgomery, C.~M. Mastrangelo, Some statistical process control methods
  for autocorrelated data, Journal of Quality Technology 23~(3) (1991)
  179--193.

\bibitem{Kasabov2003}
N.~Kasabov, S.~Pang, Transductive support vector machines and applications in
  bioinformatics for promoter recognition, in: Neural networks and signal
  processing, 2003. proceedings of the 2003 international conference on,
  Vol.~1, IEEE, 2003, pp. 1--6.

\bibitem{Raza2014c}
H.~Raza, G.~Prasad, Y.~Li, H.~Cecotti, Toward transductive learning classifiers
  for non-stationary eeg, in: Engineering in Medicine and Biology Society
  (EMBC), 2014 35th Annual International Conference of the IEEE, 2014, p.~1.

\bibitem{Raza2016_IJCNN}
H.~Raza, H.~Cecotti, G.~Prasad, A combination of transductive and inductive
  learning for handling non-stationarities in motor imagery classification, in:
  Neural Networks (IJCNN), 2016 International Joint Conference on, IEEE, 2016,
  pp. 763--770.

\bibitem{Zhu2005}
X.~Zhu, Semi-supervised learning literature survey.

\bibitem{Bishop2006}
M.~B. Christopher, Pattern Recognition and Machine Learning., Springer-Verlag
  New York, 2006.

\bibitem{Cai2008}
D.~Cai, X.~He, J.~Han, Srda: An efficient algorithm for large-scale
  discriminant analysis, IEEE transactions on knowledge and data engineering
  20~(1) (2008) 1--12.

\bibitem{Tangermann2012}
M.~Tangermann, K.-R. M{\"u}ller, A.~Aertsen, N.~Birbaumer, C.~Braun,
  C.~Brunner, R.~Leeb, C.~Mehring, K.~J. Miller, G.~Mueller-Putz, et~al.,
  Review of the bci competition iv, Frontiers in neuroscience 6 (2012) 55.

\bibitem{Kolter2003}
J.~Z. Kolter, M.~A. Maloof, Dynamic weighted majority: A new ensemble method
  for tracking concept drift, in: Data Mining, 2003. ICDM 2003. Third IEEE
  International Conference on, IEEE, 2003, pp. 123--130.

\bibitem{Lemm2005}
S.~Lemm, B.~Blankertz, G.~Curio, K.-R. Muller, Spatio-spectral filters for
  improving the classification of single trial eeg, IEEE transactions on
  biomedical engineering 52~(9) (2005) 1541--1548.

\bibitem{Ang2008}
K.~K. Ang, Z.~Y. Chin, H.~Zhang, C.~Guan, Filter bank common spatial pattern
  (fbcsp) in brain-computer interface, in: IEEE International Joint Conference
  on Neural Networks, IEEE, 2008, pp. 2390--2397.

\bibitem{Zhang2011}
H.~Zhang, Z.~Y. Chin, K.~K. Ang, C.~Guan, C.~Wang, Optimum spatio-spectral
  filtering network for brain-computer interface, IEEE Transactions on Neural
  Networks 22~(1) (2011) 52--63.

\bibitem{Gandhi2014}
V.~Gandhi, G.~Prasad, D.~Coyle, L.~Behera, T.~M. McGinnity, Quantum neural
  network-based eeg filtering for a brain-computer interface, IEEE transactions
  on neural networks and learning systems 25~(2) (2014) 278--288.

\bibitem{Gao2015}
D.~Gao, R.~Zhang, T.~Liu, F.~Li, T.~Ma, X.~Lv, P.~Li, D.~Yao, P.~Xu, Enhanced
  z-lda for small sample size training in brain-computer interface systems,
  Computational and mathematical methods in medicine 2015.

\end{thebibliography}

\end{document}